\documentclass{article}

\usepackage{microtype}
\usepackage{graphicx}
\usepackage{subfigure}
\usepackage{booktabs} 
\usepackage{xspace}
\newcommand{\eg}{e.g.\xspace}

\usepackage{xcolor}

\definecolor{aobogreen}{RGB}{0,128,0} 

\usepackage{hyperref}
\usepackage{multirow}

\usepackage[accepted]{icml2025}

\usepackage{amsmath}
\usepackage{amssymb}
\usepackage{mathtools}
\usepackage{amsthm}

\usepackage[capitalize,noabbrev]{cleveref}

\theoremstyle{plain}

\theoremstyle{definition}

\theoremstyle{remark}

\usepackage[textsize=tiny]{todonotes}

\icmltitlerunning{V-CAGE: Context-Aware Generation and Verification
for Scalable Long-Horizon Embodied Tasks}

\begin{document}

\twocolumn[
\icmltitle{V-CAGE: Context-Aware Generation and Verification \\ for Scalable Long-Horizon Embodied Tasks}



\icmlsetsymbol{equal}{*}

\begin{icmlauthorlist}
\icmlauthor{Yaru Liu}{equal,yyy}
\icmlauthor{Aobo Wang}{equal,sch}
\icmlauthor{Nanyang Ye}{comp}

\end{icmlauthorlist}

\icmlaffiliation{yyy}{Department of Computer Science and Technology, University of Cambridge, Cambridge, England. Email: yl962@cam.ac.uk}
\icmlaffiliation{comp}{Shanghai Jiao Tong University, Shanghai, China}
\icmlaffiliation{sch}{Wuhan University, Wuhan, Hubei, China}

\icmlcorrespondingauthor{Nanyang Ye}{ynylincoln@sjtu.edu.cn}


\vskip 0.3in
]



\printAffiliationsAndNotice{\icmlEqualContribution} 

\begin{abstract}
Learning long-horizon embodied behaviors from synthetic data remains challenging because generated scenes are often physically implausible, language-driven programs frequently “succeed” without satisfying task semantics, and high-level instructions require grounding into executable action sequences. To address these limitations, we introduce V-CAGE, a closed-loop framework for generating robust, semantically aligned manipulation datasets at scale. First, we propose a context-aware instantiation mechanism that enforces geometric consistency during scene synthesis. By dynamically maintaining a map of prohibited spatial areas as objects are placed, our system prevents interpenetration and ensures reachable, conflict-free configurations in cluttered environments. Second, to bridge the gap between abstract intent and low-level control, we employ a hierarchical instruction module. This decomposes high-level goals (\eg, "get ready for work") into compositional action primitives, facilitating coherent long-horizon planning. Crucially, we enforce semantic correctness through a VLM-based verification loop. Acting as a visual critic, the VLM performs rigorous rejection sampling after each subtask, filtering out "silent failures" where code executes but fails to achieve the visual goal. Experiments demonstrate that V-CAGE yields datasets with superior physical and semantic fidelity, significantly boosting the success rate and generalization of downstream policies compared to non-verified baselines.
\end{abstract}

\section{Introduction}
\label{introduction}
Recent advancements in Large Language Models (LLMs) and Vision-Language Models (VLMs) have demonstrated remarkable capabilities in reasoning and planning~\cite{palme, saycan}. However, translating these high-level semantic capabilities into robust embodied agents capable of executing long-horizon manipulation tasks remains a formidable challenge. While the "Scaling Laws" have proven effective in NLP and Vision, applying them to robotics is hindered by the lack of high-quality, large-scale demonstration data. Collecting real-world trajectories is expensive, unsafe, and difficult to scale~\cite{rt1}. Therefore, learning from synthetic data has emerged as a promising paradigm~\cite{MimicGen, wang2024robogen}.

Despite its promise, naively generated synthetic data often suffers from a significant "quality gap" that severely limits downstream policy performance. This gap is particularly pronounced in long-horizon tasks, for example "prepare the desk for work", where success depends on the precise execution of sequential subtasks. We identify two fundamental failure modes in existing data generation pipelines. 
First, Geometric Inconsistency: Standard procedural generation often places objects without considering the dynamic evolution of the workspace. As the scene becomes cluttered, objects are instantiated in conflicting poses or unstable configurations, leading to physics simulator crashes or unrealistic interpenetrations. 
Second, Semantic Misalignment: Language-conditioned code generation can be brittle. A generated script might execute without runtime errors, but fail to achieve the intended semantic goal (\eg, the switch was not actually toggled, or the object was placed off the pad). Training on such "false positive" data introduces noise that catastrophic degrades the agent's ability to reason about cause and effect~\cite{causalconfusion}.

To address these challenges, we present V-CAGE (VLM-Guided Context-Aware Generation for Embodied Planning), a closed-loop framework designed to synthesize long-horizon manipulation trajectories. Unlike open-loop generation methods that blindly execute LLM-produced plans, V-CAGE integrates geometric constraints and visual verification directly into the generation pipeline. 
Our core insight treats data synthesis as a rigorous optimization process, aggressively pruning invalid trajectories to ensure only high-fidelity data reaches the training buffer.

V-CAGE operates on three hierarchical levels. At the planning level, we leverage an LLM, for example Pangu \cite{chen2025pangu}, to ground abstract user instructions into executable action sequences. At the geometric level, we propose a context-aware instantiation mechanism. By maintaining a dynamic map of "prohibited volumes" that updates after each object placement, we ensure that new objects are instantiated only in feasible, collision-free regions, effectively solving the packing problem in cluttered scenes. Crucially, at the verification level, we employ a VLM, Gemini3 \cite{gemini3report2025}, as a semantic critic. Treating data generation as a rejection sampling problem, the VLM evaluates the visual outcome of each subtask. Trajectories that fail to meet visual success criteria are rejected and regenerated, ensuring that the final dataset consists exclusively of physically plausible and semantically correct demonstrations.

Our main contributions are summarized as follows: 
(1) We introduce V-CAGE, a scalable, closed-loop pipeline for synthesizing high-fidelity trajectory data for long-horizon tasks, integrating geometric constraints with visual verification. 
(2) We propose a context-aware instantiation mechanism that maintains an evolving map of prohibited volumes during scene generation, ensuring valid object placement in cluttered environments. 
(3) We formulate data validation as a VLM-guided rejection sampling process, using visual critics to filter out "silent failures" where code executes without satisfying task semantics. 
(4) We demonstrate that policies trained on V-CAGE data significantly outperform baselines in terms of success rate and generalization across diverse, cluttered scenarios.


\section{Related work}
\label{relatedwork}

\subsection{LLM-Generated Robotic Simulation Tasks and Training Data}

A growing line of work leverages LLMs to automatically construct robotic simulation tasks and training data, aiming to reduce the substantial human effort required for task design and data collection. \citet{gensim} introduce an LLM-driven code generation pipeline that writes simulation environments, task specifications, and expert policies, enabling the automatic creation of rich task libraries for manipulation in synthetic environments. By combining goal-directed and exploratory task generation modes, GenSim scales from hand-designed benchmarks to a much larger set of diverse tasks and shows improved task-level generalization and sim-to-real transfer for learned policies.

Building on this idea, \citet{gensim2} propose a more scalable framework that exploits coding LLMs with multi-modal and reasoning capabilities to create complex and realistic simulation tasks, including long-horizon manipulation with articulated objects. The framework first uses an LLM to propose tasks and generate executable task code, then employs planning and reinforcement learning solvers to automatically produce demonstrations at scale, and finally trains a language-conditioned policy architecture on the generated dataset. This pipeline can generate data for up to hundreds of articulated-object tasks and achieves strong zero-shot sim-to-real transfer and performance gains when co-trained with limited real-world data \citep{gensim2}.

In parallel, \citet{wang2024robogen} treat foundation and generative models as generative simulators rather than direct controllers. They define a self-guided robotic agent that autonomously proposes new tasks, generates corresponding environments, and acquires diverse skills via generative simulation. The system continuously expands a training corpus that covers over a hundred tasks and demonstrates that such procedurally generated, model-driven simulations can rival or outperform human-authored datasets for multi-task robot learning.

For bimanual manipulation and digital-twin scenarios, \citet{robotwin} and \citet{robotwin2} integrate LLMs into a generative digital-twin pipeline. Starting from single RGB images, RoboTwin \citep{robotwin} uses 3D generative models to create diverse object instances and employs LLM agents to synthesize task programs for dual-arm manipulation, yielding an aligned synthetic–real benchmark and scalable expert data generation. RoboTwin 2.0 \citep{robotwin2} further extends this framework into a large-scale data generator and benchmark with over fifty tasks and hundreds of object categories, incorporating multimodal LLM-based program synthesis and extensive domain randomization to improve robustness of learned bimanual policies.

Beyond fully LLM-coded pipelines, benchmark-focused efforts such as RoboCAS \citep{robocas} focuses on complex object arrangement scenarios in robotic manipulation. RoboCAS defines a benchmark for long-horizon manipulation in cluttered arrangement scenes and uses flexible scripted policies to collect diverse demonstrations across scattered, ordered, and stacked configurations. It highlights the importance of compositional scene and task complexity for evaluating generalist manipulation policies and foundation models.

Overall, existing work shows that LLMs can significantly reduce human effort in designing simulation tasks and generating demonstrations, while improving task diversity, long-horizon complexity, and sim-to-real transfer. 

\subsection{LLMs for Long-Horizon Robotic Task Planning}

A complementary line of research employs LLMs as high-level planners for long-horizon robotic tasks. PaLM-SayCan \citep{Ahn22SayCan} grounds a large language model in a discrete set of pre-trained skills, combining the LLM’s estimate of which skill is useful with value functions that predict whether a skill will succeed in the current state. This affordance-grounded formulation enables a mobile manipulator to execute abstract multi-step instructions in real environments while maintaining feasibility and safety. DELTA \citep{Liu24DELTA} uses 3D scene graphs as structured environment representations and leverages LLMs to (i) generate formal planning domain and problem specifications and (ii) decompose long-term goals into a sequence of sub-goals, which are then solved autoregressively by a classical task planner. By coupling scene-graph reasoning with LLM-based decomposition, DELTA achieves higher success rates and lower planning time on large-scale, long-term navigation and manipulation tasks.

More recently, RoboHorizon \citep{Chen25RoboHorizon} proposes an LLM-assisted multi-view world model for long-horizon manipulation, instantiated within a Recognize--Sense--Plan--Act pipeline. In RoboHorizon, an LLM generates dense reward structures over multi-stage sub-tasks from language instructions, and keyframe discovery is integrated into a visual world model to better capture critical steps in long-horizon processes, leading to substantial improvements over prior model-based RL baselines on RLBench and FurnitureBench. Other works explore LLMs for translating natural-language task descriptions into formal constraints or planning specifications for task and motion planning (TAMP) \citep{Guo25CaStL}, further extending LLM-based planning to larger action spaces and more complex constraints. 

Beyond robotics, \citet{Meyerson25MAKER} propose MAKER, an instance of a massively decomposed agentic process (MDAP), which solves a Towers-of-Hanoi task with over one million LLM steps and zero errors by combining extreme subtask decomposition, subtask-level voting, and red-flag-based error filtering. Their results demonstrate that scaling long-horizon LLM execution can come not only from stronger base models, but also from systematic decomposition and fine-grained error correction at the agentic level.

These approaches largely position LLMs as high-level planners that output symbolic sub-goals, skill sequences, or formal planning problems to be executed by separate low-level controllers. 

\subsection{VLM-Driven Manipulation and Planning}
Recent works have significantly advanced the integration of LLMs and VLMs into robotic decision-making. VILA~\cite{vila} demonstrates the efficacy of pre-training VLMs on massive video-text data to unlock strong reasoning capabilities for long-horizon planning tasks. Similarly, OmniManip~\cite{omnimanip} leverages large-scale datasets to train generalist policies capable of performing diverse manipulation skills across varied environments. To address the grounding problem in these high-level plans, MOKA~\cite{moka} introduces a mark-based visual prompting framework, utilizing VLMs to generate affine transformations and grasp poses by ``marking'' keypoints on target objects.
However, these approaches primarily focus on inference-time reasoning or policy learning assuming the existence of data. They do not explicitly address the scalability bottleneck of acquiring high-quality, long-horizon demonstration data from scratch. While VILA and MOKA improve how a robot acts, V-CAGE focuses on synthesizing the training data itself, using VLMs not just as planners, but as rigorous automated verifiers.

\subsection{Scalable Synthetic Data Generation}
To overcome the scarcity of real-world data, generative simulation has emerged as a promising paradigm. GenManip~\cite{genmanip} pioneered the use of LLMs to automatically generate simulation environments and task codes, scaling data collection beyond manual engineering. These pipelines typically follow an open-loop ``generate-and-execute'' protocol.
In contrast, V-CAGE identifies a critical limitation in prior generative works: the lack of closed-loop verification and spatial awareness. Naive generation often results in ``silent failures''—where generated code executes without errors but fails to achieve the semantic goal—or geometric conflicts in cluttered scenes. V-CAGE advances this domain by introducing a context-aware instantiation mechanism to handle dynamic spatial constraints and a VLM-guided rejection sampling loop. This ensures that every trajectory added to the dataset is not only generated but strictly validated for physical and semantic correctness.

\begin{figure*}
  \centering
   \includegraphics[width=\linewidth]{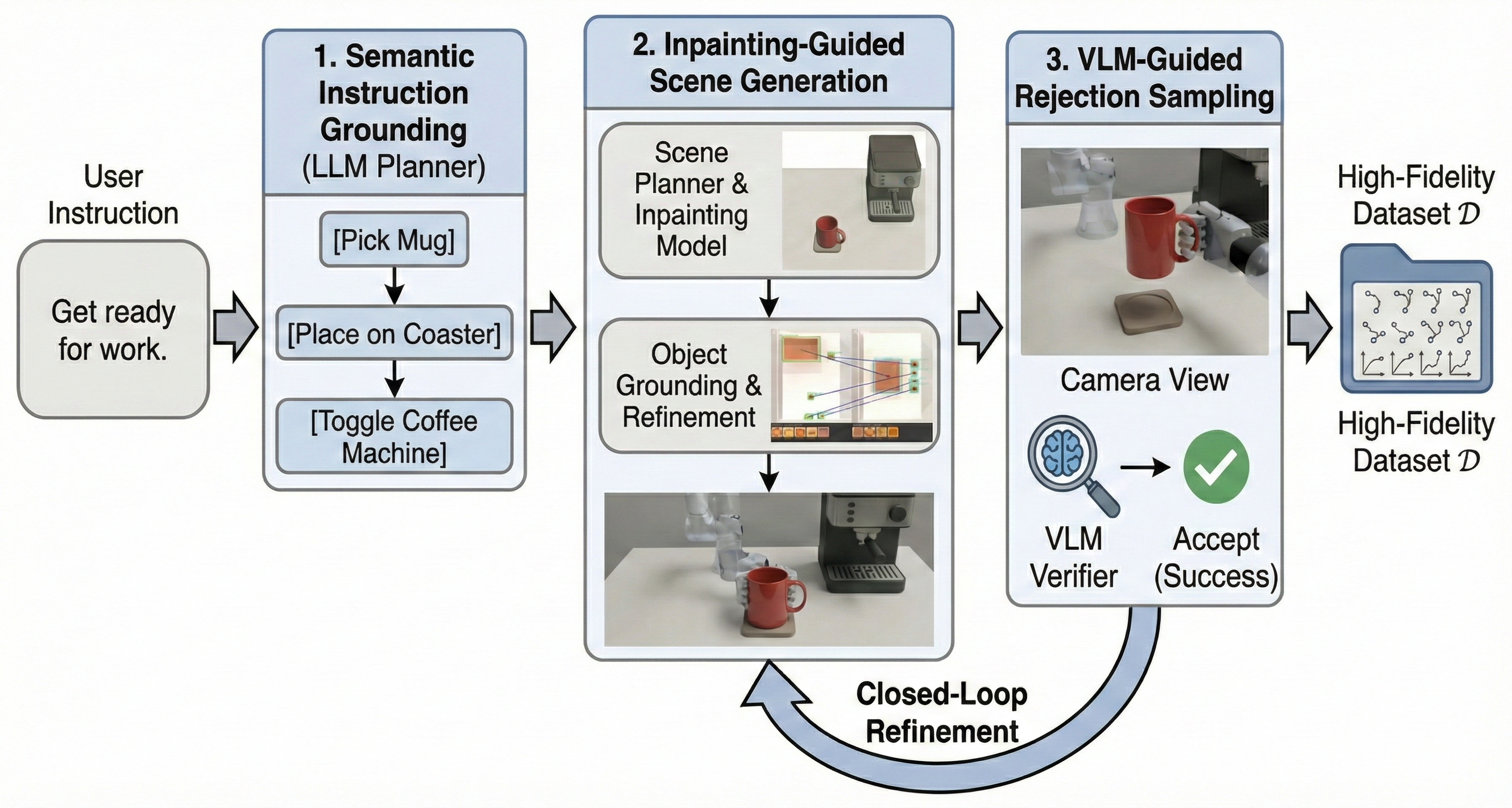}
    \caption{Overview of the V-CAGE framework. A high-level instruction is decomposed by an LLM into subtasks. During simulation, the VLM Gemini3 \cite{gemini3report2025} acts as a visual critic after each step, verifying semantic success based on the post-execution image. If a failure is detected, the trajectory generation is aborted to ensure high-fidelity data.}  
    \label{fig:exp-result-mixed}
\end{figure*}

\section{Method}
\label{sec:method}

We introduce V-CAGE (VLM-Guided Context-Aware Generation for Embodied Planning), a closed-loop framework for synthesizing high-fidelity manipulation datasets. The pipeline consists of three hierarchical modules: (1) Semantic Instruction Grounding via LLMs, (2) Context-Aware Scene Instantiation to resolve geometric constraints, and (3) VLM-Guided Rejection Sampling to ensure causal task success.

\subsection{Problem Formulation}
Our goal is to generate a dataset of successful long-horizon trajectories $\mathcal{D} = \{ \tau_i \}_{i=1}^N$ conditioned on high-level user instructions $I_{user}$. A trajectory $\tau$ consists of a sequence of state-action pairs satisfying a sequence of subgoals.
The core challenge is that the joint probability of a successful trajectory $P(\tau | I_{user})$ factorizes into both physical feasibility $P_{phys}(\tau)$ and semantic correctness $P_{sem}(\tau)$. Naive generation maximizes neither; V-CAGE explicitly optimizes both via geometric and visual constraints.

\subsection{Semantic Instruction Grounding}
Given an abstract instruction $I_{user}$ (e.g., ``Get ready for work''), we employ a Language Model, Pangu \cite{chen2025pangu}, as a high-level planner to decompose the intent into a sequence of executable subtasks $\mathcal{T} = \{T_1, T_2, \dots, T_k\}$.
Each subtask $T_i$ is mapped to a parameterized motion primitive (e.g., \texttt{turn\_switch}, \texttt{place\_object}). This decomposition transforms the long-horizon problem into a series of short-horizon grounding problems.



\subsection{VLM-Guided Rejection Sampling}
Even with valid geometry, execution artifacts (e.g., gripper slip, switch not toggled) can lead to semantic failures. To filter these ``silent failures,'' we model data generation as a Rejection Sampling process using a Vision-Language Model, Gemini3 \cite{gemini3report2025}, as the density estimator.

Let $\phi_{VLM}(I_{img}, T_i) \to \{0, 1\}$ be the verification function, where $I_{img}$ is the post-execution image and $T_i$ is the subtask description. The VLM acts as a visual critic. For a sequence of tasks $T_{1:k}$, a trajectory is accepted into the dataset only if:
\begin{equation}
    \prod_{i=1}^{k} \phi_{VLM}(I_{img}^{(i)}, T_i) = 1
\end{equation}
If $\phi_{VLM}(\cdot) = 0$ at any step, the generation episode is aborted, and the scene is reset (or re-sampled). This rigorous filtering ensures that the final dataset $\mathcal{D}$ contains only trajectories that are effectively causally valid, bridging the gap between code execution and visual reality.

\subsection{Generative Scene and Task Synthesis Pipeline}
\label{sec:scene_generation}

\begin{figure*}[t]
    \centering
    \includegraphics[width=\linewidth]{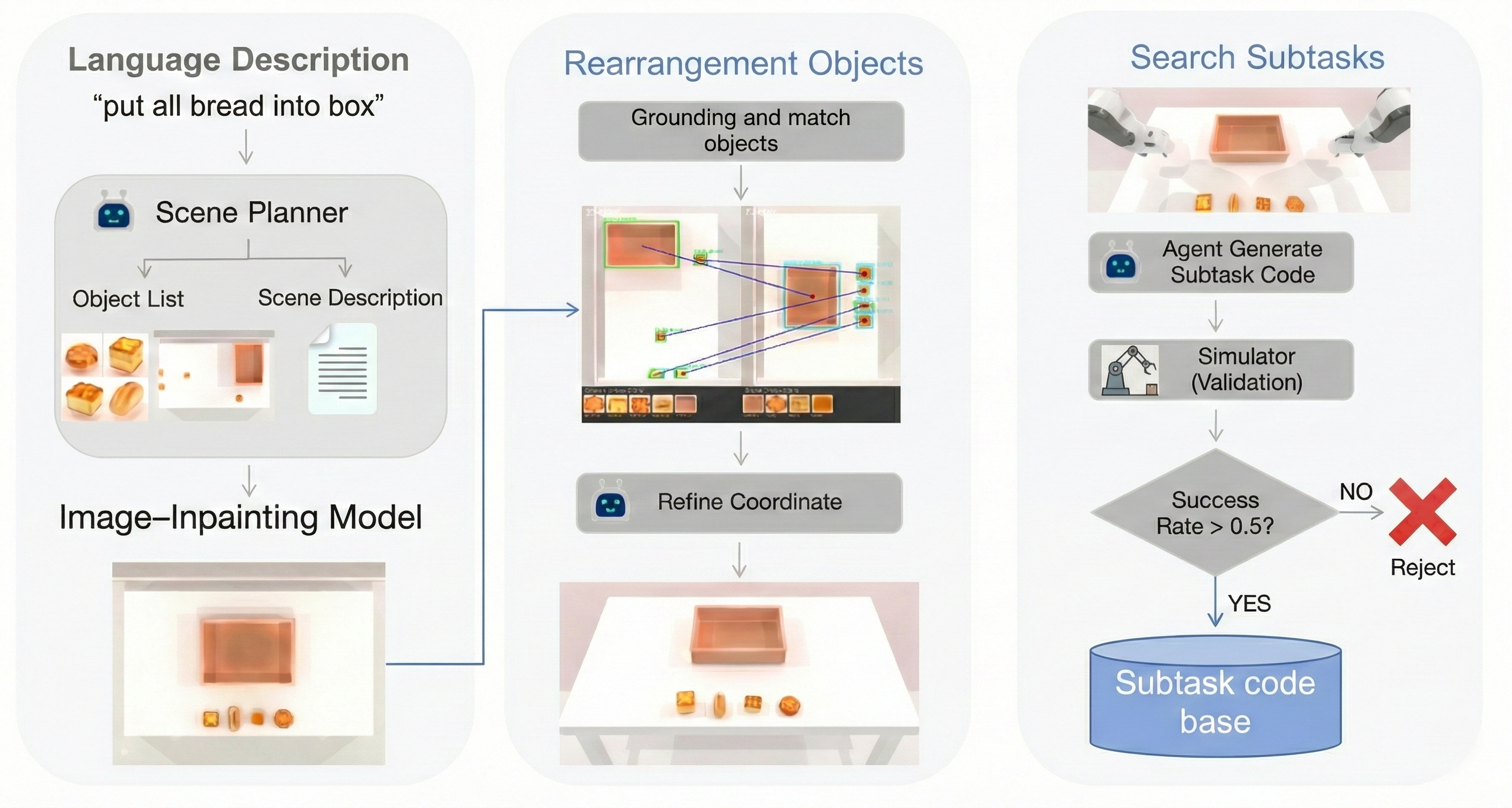}
    \caption{The proposed pipeline for autonomous scene generation and task verification. It bridges high-level semantic planning with low-level physical simulation through visual rearrangement and iterative refinement.}
    \label{fig:pipeline}
\end{figure*}

To bridge the gap between abstract natural language instructions and executable robotic skills, we propose a five-stage generative pipeline. This framework synthesizes physically valid, semantically rich scenes and validates corresponding manipulation subtasks within a simulation environment.

The process initiates with \textbf{Language-Guided Asset Retrieval and Scene Planning}. Given a high-level user prompt (e.g., ``organize the breakfast table''), a Large Language Model (LLM) acts as a semantic planner. It parses the instruction to retrieve a set of compatible object assets $\mathcal{O} = \{o_1, \ldots, o_n\}$ from our annotated database, utilizing metadata such as object semantics and grasp affordances. Simultaneously, the LLM generates a detailed natural language scene description $\mathcal{D}_{scene}$ that explicitly specifies spatial constraints and layout logic (e.g., ``place the bread inside the wooden box'').

To realize this description visually, we employ a \textbf{Visual-Conditioned Rearrangement} strategy. The selected objects are first instantiated in the simulation via a randomized collision-free scattering to produce an initial top-down orthographic view $I_{init}$. This image, along with the generated description $\mathcal{D}_{scene}$, serves as input to an image-inpainting model (NanoBanana-Pro). The model hallucinates a goal configuration $I_{goal} = \mathcal{M}(I_{init}, \mathcal{D}_{scene})$ that respects the semantic spatial relations described in the text while maintaining visual consistency with the assets.

Subsequently, we perform \textbf{Scene Refinement and Grounding}. We extract the 2D pixel coordinates of objects from $I_{goal}$ via template matching against $I_{init}$ and project them back into the 3D simulation space. To ensure physical plausibility, these raw coordinates are passed to a Refine Agent. The agent optimizes the final pose configuration $\mathbf{P}^*$ by minimizing a cost function that penalizes object collisions and deviations from the semantic description:
\begin{equation}
    \mathbf{P}^* = \mathop{\arg\min}_{\mathbf{P}} \left( \mathcal{L}_{collision}(\mathbf{P}) + \lambda \mathcal{L}_{semantic}(\mathbf{P}, \mathcal{D}_{scene}) \right)
\end{equation}
where $\mathbf{P}$ represents the set of 6-DoF poses for all objects. This step corrects artifacts introduced by the image generation process, ensuring the scene is physically stable.

Upon stabilizing the scene, the system proceeds to \textbf{Subtask Code Generation}. We define a library of parameterized skill templates (e.g., \texttt{pick\_and\_place(obj, container)}). The LLM identifies actionable relationships within the scene---specifically pairing objects with valid grasp annotations to containers with placement annotations---and instantiates executable Python code for each potential subtask $T_i$.

Finally, we enforce a \textbf{Simulation-Based Verification} mechanism. Each generated subtask $T_i$ is executed in the physics simulator. We measure the success rate $SR(T_i)$ over multiple trials. A subtask is deemed valid and committed to the subtask database only if it satisfies a robustness threshold $\tau$:
\begin{equation}
    \text{Status}(T_i) = 
    \begin{cases} 
    \text{Accept} & \text{if } SR(T_i) > 0.5 \\
    \text{Reject} & \text{otherwise}
    \end{cases}
\end{equation}
This closed-loop verification ensures that the generated data is not only visually plausible but functionally actionable for downstream policy learning.

\begin{figure}
    \centering
    \includegraphics[width=\linewidth]{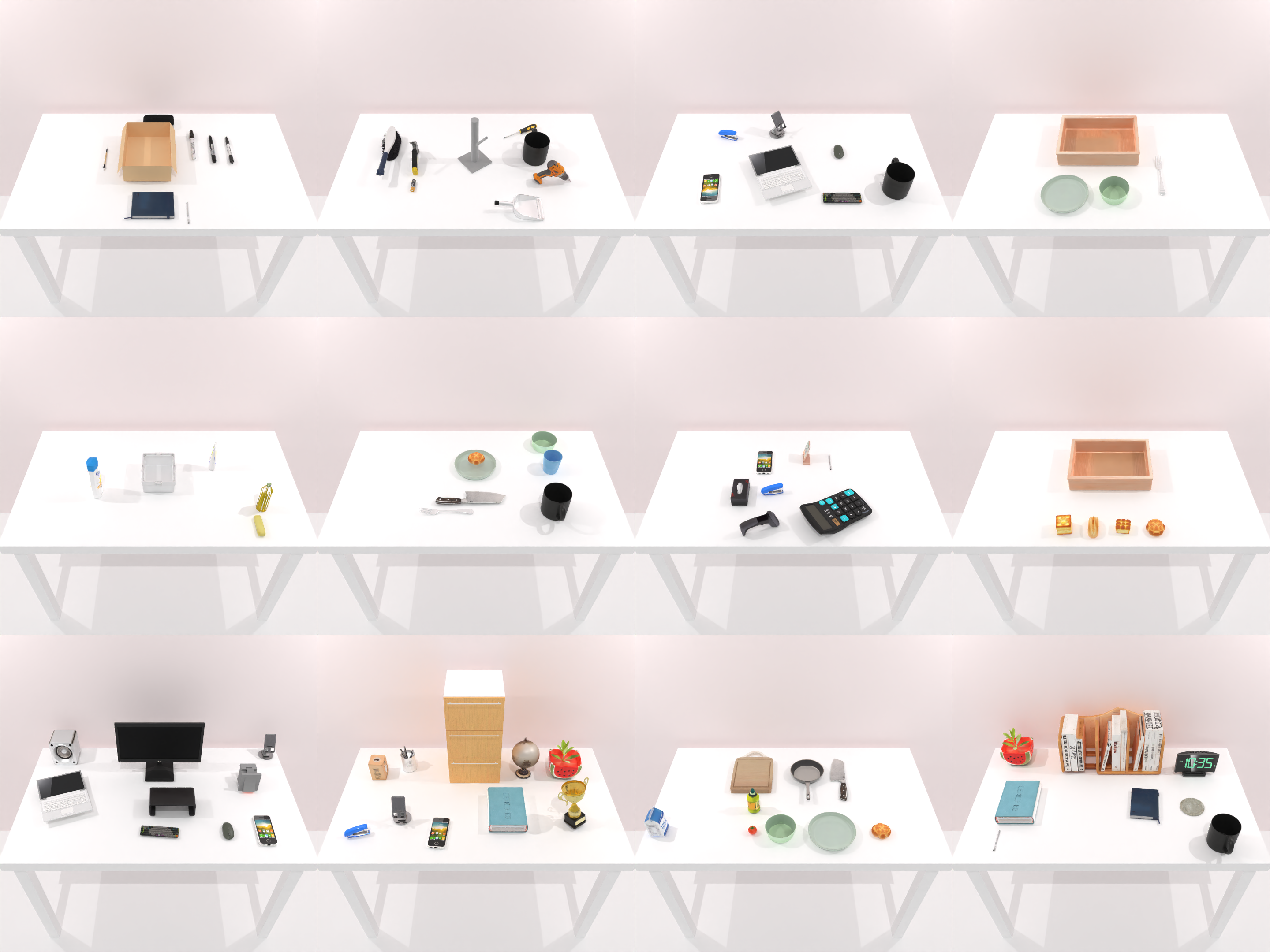}
    \caption{Example scenes generated by V-CAGE}
    \label{fig:scenes}
\end{figure}

\section{Experiments}
\label{sec:experiments}

In this section, we empirically evaluate the effectiveness of the V-CAGE framework. Our experiments are designed to answer two primary questions: (1) Does incorporating context-aware geometric constraints and VLM-based verification improve the quality of synthetic training data? (2) How does this data quality translate to the performance of downstream policies in diverse, long-horizon manipulation tasks?

\subsection{Experimental Setup}
We evaluate our method on a suite of 35 diverse manipulation tasks derived from the RoboTwin benchmark~\cite{robotwin}. These tasks range from simple object rearrangement to complex, multi-stage activities such as table setting and semantic sorting in cluttered environments. 
We compare our full pipeline (\textbf{V-CAGE}) against a strong baseline (\textbf{Vanilla}), which employs the same LLM planner and code generation module but lacks the closed-loop VLM verification and dynamic context-aware instantiation. In the Vanilla setting, success is determined solely by the absence of execution errors (e.g., Python exceptions or collision crashes) during simulation, a standard proxy used in prior open-loop generation works.

\subsection{Main Results}
\label{subsec:main_results}

We trained diffusion-based manipulation policies using datasets generated by both V-CAGE and the Vanilla baseline. We evaluate the policy performance using three metrics: \textbf{Average Success Rate} across all tasks, and the \textbf{Top-K Success Rate} (for $K=5$ and $10$), which measures the reliability of the best-performing checkpoints.

The quantitative results are summarized in Table~\ref{tab:performance_comparison}. V-CAGE achieves a significant improvement over the baseline across all metrics. Notably, the \textbf{Average Success Rate} increases from 46.86\% to 64.58\%, a substantial gain of +17.72\%. This performance gap highlights the superior quality of the data synthesized by our framework.

\begin{table}[h]
    \centering
    \caption{\textbf{Performance Comparison on Downstream Manipulation Tasks.} We report the success rates averaged over 35 long-horizon tasks from the RoboTwin benchmark. The "Vanilla" baseline relies on open-loop generation, while "V-CAGE" (Ours) integrates VLM-based semantic verification. The gap in Top-10 performance highlights the stability provided by our method.}
    \label{tab:performance_comparison}
    \vskip 0.15in
    \begin{small}
    \begin{sc}
    \begin{tabular}{lcc}
        \toprule
        \textbf{Metric} & \textbf{V-CAGE (Ours)} & \textbf{Vanilla} \\
        \midrule
        Average Success Rate & \textbf{64.58\%} & 46.86\% \\
        Top 5 Success Rate & \textbf{100.00\%} & 92.00\% \\
        Top 10 Success Rate & \textbf{100.00\%} & 77.00\% \\
        \bottomrule
    \end{tabular}
    \end{sc}
    \end{small}
    \vskip -0.1in
\end{table}

\subsection{The Critical Role of VLM Verification}
The most revealing insight from Table~\ref{tab:performance_comparison} lies in the disparity between the Top-K metrics. 
While the Vanilla baseline achieves a respectable 92\% in the Top-5 metric, its performance drops precipitously to 77\% for the Top-10. In stark contrast, V-CAGE maintains a \textbf{100\% success rate} for both Top-5 and Top-10.

This stability attests to the critical role of the VLM as a semantic filter. In the Vanilla pipeline, trajectories are labeled as "successful" as long as the generated code executes without throwing system exceptions. However, this introduces a high volume of \textit{silent failures}—instances where the robot moves and the code terminates gracefully, but the semantic goal is not achieved (e.g., an object slips from the gripper, or a switch is missed by millimeters). These false positives introduce severe label noise into the training dataset, causing the downstream policy to learn incorrect associations between actions and outcomes.

By integrating the VLM verification loop, V-CAGE effectively performs rejection sampling on the \textit{semantic} outcome rather than just the \textit{programmatic} outcome. The VLM acts as a rigorous visual critic, pruning these silent failures before they enter the training buffer. The fact that our Top-10 performance remains perfect (100\%) indicates that the VLM ensures a much higher "purity" of valid demonstrations, allowing the policy to learn robust, generalizable behaviors without being confused by noisy supervision. Consequently, VLM-based verification is not merely an auxiliary feature but a fundamental prerequisite for scaling up synthetic data generation for long-horizon tasks.

\section{Conclusion}
In this work, we presented V-CAGE, a closed-loop framework for synthesizing high-fidelity, long-horizon manipulation datasets. Our context-aware instantiation mechanism eliminates physical conflicts in complex, multi-object scenes, while our VLM-guided rejection sampling pipeline rigorously filters out execution failures that plague unverified synthetic data.
Experiments demonstrate the effectiveness of V-CAGE in improving policy robustness to cluttered environments, enabling generalization to unseen tasks. These findings underscore the critical importance of data quality over mere quantity in the era of generative robotics.

\textbf{Limitations and Future Work.} While V-CAGE ensures high data fidelity, the rejection sampling process can be computationally expensive for highly complex tasks with low inherent success probabilities. Future work could explore using the VLM critic to provide dense reward signals for correcting failed trajectories in-the-loop, rather than discarding them, thereby improving generation efficiency. Additionally, extending V-CAGE to dynamic interactions involving deformable objects or fluids remains an exciting direction for future research.

\nocite{langley00}

\bibliography{example_paper}

\begin{thebibliography}{23}
\providecommand{\natexlab}[1]{#1}
\providecommand{\url}[1]{\texttt{#1}}
\expandafter\ifx\csname urlstyle\endcsname\relax
  \providecommand{\doi}[1]{doi: #1}\else
  \providecommand{\doi}{doi: \begingroup \urlstyle{rm}\Url}\fi

\bibitem[Ahn et~al.(2022{\natexlab{a}})Ahn, Brohan, Brown, Chebotar, Cortes, David, Finn, Fu, Gopalakrishnan, Hausman, et~al.]{Ahn22SayCan}
Ahn, M., Brohan, A., Brown, N., Chebotar, Y., Cortes, O., David, B., Finn, C., Fu, C., Gopalakrishnan, K., Hausman, K., et~al.
\newblock Do as i can, not as i say: Grounding language in robotic affordances.
\newblock \emph{arXiv preprint arXiv:2204.01691}, 2022{\natexlab{a}}.

\bibitem[Ahn et~al.(2022{\natexlab{b}})Ahn, Brohan, Brown, Chebotar, Cortes, et~al.]{saycan}
Ahn, M., Brohan, A., Brown, N., Chebotar, Y., Cortes, O., et~al.
\newblock Do as i can, not as i say: Grounding language in robotic affordances.
\newblock In \emph{Conference on Robot Learning (CoRL)}, 2022{\natexlab{b}}.

\bibitem[Brohan et~al.(2023)Brohan, Brown, Carbajal, Chebotar, Dabis, et~al.]{rt1}
Brohan, A., Brown, N., Carbajal, J., Chebotar, Y., Dabis, J., et~al.
\newblock Rt-1: Robotics transformer for real-world control at scale.
\newblock In \emph{Robotics: Science and Systems (RSS)}, 2023.

\bibitem[Chen et~al.(2025{\natexlab{a}})Chen, Wang, Han, Li, Li, Bi, Li, Wang, Mi, Zhu, Wang, Song, Nie, Wu, He, Hu, Tang, Tao, Chen, and Wang]{chen2025pangu}
Chen, H., Wang, Y., Han, K., Li, D., Li, L., Bi, Z., Li, J., Wang, H., Mi, F., Zhu, M., Wang, B., Song, K., Nie, Y., Wu, X., He, W., Hu, H., Tang, Y., Tao, D., Chen, X., and Wang, Y.
\newblock Pangu embedded: An efficient dual-system llm reasoner with metacognition.
\newblock \emph{arXiv preprint arXiv:2505.22375}, 2025{\natexlab{a}}.
\newblock URL \url{https://arxiv.org/abs/2505.22375}.

\bibitem[Chen et~al.(2025{\natexlab{b}})Chen, Chen, Chen, Cai, Liu, Li, Liang, Lin, Ge, Gu, Guo, Nian, Xie, Chen, Su, Xu, Liu, Hu, Gao, Wang, Liang, Qin, Yang, Luo, and Mu]{robotwin2}
Chen, T., Chen, Z., Chen, B., Cai, Z., Liu, Y., Li, Z., Liang, Q., Lin, X., Ge, Y., Gu, Z., Guo, Y., Nian, T., Xie, X., Chen, Q., Su, K., Xu, T., Liu, G., Hu, M., Gao, H., Wang, K., Liang, Z., Qin, Y., Yang, X., Luo, P., and Mu, Y.
\newblock {RoboTwin} 2.0: A scalable data generator and benchmark with strong domain randomization for robust bimanual robotic manipulation.
\newblock \emph{arXiv preprint arXiv:2506.18088}, 2025{\natexlab{b}}.

\bibitem[Chen et~al.(2025{\natexlab{c}})Chen, Huo, Chen, and Gao]{Chen25RoboHorizon}
Chen, Z., Huo, J., Chen, Y., and Gao, Y.
\newblock Robohorizon: An {LLM}-assisted multi-view world model for long-horizon robotic manipulation.
\newblock \emph{arXiv preprint arXiv:2501.06605}, 2025{\natexlab{c}}.

\bibitem[de~Haan et~al.(2019)de~Haan, Jayaraman, and Levine]{causalconfusion}
de~Haan, P., Jayaraman, D., and Levine, S.
\newblock Causal confusion in imitation learning.
\newblock In \emph{Advances in Neural Information Processing Systems (NeurIPS)}, 2019.

\bibitem[Driess et~al.(2023)Driess, Xia, Sajjadi, Lynch, Chowdhery, et~al.]{palme}
Driess, D., Xia, F., Sajjadi, M.~S., Lynch, C., Chowdhery, A., et~al.
\newblock Palm-e: An embodied multimodal language model.
\newblock In \emph{International Conference on Machine Learning (ICML)}, 2023.

\bibitem[{Gemini Team, Google}(2025)]{gemini3report2025}
{Gemini Team, Google}.
\newblock Gemini 3: Frontier multimodal intelligence.
\newblock \emph{arXiv preprint}, 2025.
\newblock URL \url{https://deepmind.google/technologies/gemini/}.
\newblock Technical Report.

\bibitem[Guo et~al.(2025)Guo, Kingston, and Kavraki]{Guo25CaStL}
Guo, W., Kingston, Z., and Kavraki, L.~E.
\newblock {CaStL}: Constraints as specifications through {LLM} translation for long-horizon task and motion planning.
\newblock In \emph{Proceedings of the IEEE International Conference on Robotics and Automation (ICRA)}, pp.\  11957--11964, 2025.
\newblock \doi{10.1109/ICRA55743.2025.11127555}.

\bibitem[Hua et~al.(2024)Hua, Liu, Macaluso, Lin, Zhang, Xu, and Wang]{gensim2}
Hua, P., Liu, M., Macaluso, A., Lin, Y., Zhang, W., Xu, H., and Wang, L.
\newblock Gensim2: Scaling robot data generation with multi-modal and reasoning {LLM}s.
\newblock In \emph{Proceedings of the Conference on Robot Learning (CoRL)}, 2024.
\newblock CoRL 2024.

\bibitem[Langley(2000)]{langley00}
Langley, P.
\newblock Crafting papers on machine learning.
\newblock In Langley, P. (ed.), \emph{Proceedings of the 17th International Conference on Machine Learning (ICML 2000)}, pp.\  1207--1216, Stanford, CA, 2000. Morgan Kaufmann.

\bibitem[Lin et~al.(2024)Lin, Yin, Ping, Lu, Molchanov, Han, and Alvarez]{vila}
Lin, J., Yin, H., Ping, W., Lu, Y., Molchanov, Pavlo~andlob, A., Han, S., and Alvarez, J.~M.
\newblock Vila: On pre-training for visual language models.
\newblock In \emph{Proceedings of the IEEE/CVF Conference on Computer Vision and Pattern Recognition}, 2024.

\bibitem[Liu et~al.(2024{\natexlab{a}})Liu, Lin, Yan, Yi, Abbeel, and Gao]{moka}
Liu, F., Lin, K., Yan, H., Yi, L., Abbeel, P., and Gao, Y.
\newblock Moka: Open-vocabulary robotic manipulation through mark-based visual prompting.
\newblock \emph{arXiv preprint arXiv:2403.03174}, 2024{\natexlab{a}}.

\bibitem[Liu et~al.(2024{\natexlab{b}})Liu, Palmieri, Koch, Georgievski, and Aiello]{Liu24DELTA}
Liu, Y., Palmieri, L., Koch, S., Georgievski, I., and Aiello, M.
\newblock {DELTA}: Decomposed efficient long-term robot task planning using large language models.
\newblock \emph{arXiv preprint arXiv:2404.03275}, 2024{\natexlab{b}}.

\bibitem[Mandlekar et~al.(2023)Mandlekar, Nasiriany, Wen, Akinola, and Zhu]{MimicGen}
Mandlekar, A., Nasiriany, S., Wen, B., Akinola, I., and Zhu, Y.
\newblock Mimicgen: A data generation system for scalable robot learning using human demonstrations.
\newblock In \emph{Conference on Robot Learning (CoRL)}, 2023.

\bibitem[Meyerson et~al.(2025)Meyerson, Paolo, Dailey, Shahrzad, Francon, Hayes, Qiu, Hodjat, and Miikkulainen]{Meyerson25MAKER}
Meyerson, E., Paolo, G., Dailey, R., Shahrzad, H., Francon, O., Hayes, C.~F., Qiu, X., Hodjat, B., and Miikkulainen, R.
\newblock Solving a million-step {LLM} task with zero errors.
\newblock \emph{arXiv preprint arXiv:2511.09030}, 2025.

\bibitem[Mu et~al.(2025)Mu, Chen, Chen, Peng, Lan, Gao, Liang, Yu, Zou, Xu, Lin, Xie, Ding, and Luo]{robotwin}
Mu, Y., Chen, T., Chen, Z., Peng, S., Lan, Z., Gao, Z., Liang, Z., Yu, Q., Zou, Y., Xu, M., Lin, L., Xie, Z., Ding, M., and Luo, P.
\newblock {RoboTwin}: Dual-arm robot benchmark with generative digital twins.
\newblock In \emph{Proceedings of the Computer Vision and Pattern Recognition Conference (CVPR)}, pp.\  27649--27660, June 2025.

\bibitem[{Team Octo} et~al.(2023){Team Octo}, Ghosh, Walke, Pertsch, Black, Mees, Quureshi, He, et~al.]{omnimanip}
{Team Octo}, Ghosh, D., Walke, H., Pertsch, K., Black, K., Mees, O., Quureshi, S., He, T., et~al.
\newblock Octo: An open-source generalist robot policy.
\newblock \emph{arXiv preprint arXiv:2305.10455}, 2023.

\bibitem[Wang et~al.(2024{\natexlab{a}})Wang, Ling, Yuan, Shridhar, Bao, Qin, Wang, Xu, and Wang]{gensim}
Wang, L., Ling, Y., Yuan, Z., Shridhar, M., Bao, C., Qin, Y., Wang, B., Xu, H., and Wang, X.
\newblock Gensim: Generating robotic simulation tasks via large language models.
\newblock In \emph{Proceedings of the International Conference on Learning Representations (ICLR)}, 2024{\natexlab{a}}.
\newblock ICLR 2024.

\bibitem[Wang et~al.(2024{\natexlab{b}})Wang, Zhao, Xu, Feng, Jiang, Xu, Wong, and Lim]{genmanip}
Wang, L., Zhao, Y., Xu, Y., Feng, T., Jiang, Y., Xu, D., Wong, J., and Lim, J.~J.
\newblock Gensim: Generating robotic simulation tasks via large language models.
\newblock In \emph{International Conference on Learning Representations (ICLR)}, 2024{\natexlab{b}}.

\bibitem[Wang et~al.(2024{\natexlab{c}})Wang, Xian, Chen, Wang, Wang, Fragkiadaki, Erickson, Held, and Gan]{wang2024robogen}
Wang, Y., Xian, Z., Chen, F., Wang, T.-H., Wang, Y., Fragkiadaki, K., Erickson, Z., Held, D., and Gan, C.
\newblock Robogen: Towards unleashing infinite data for automated robot learning via generative simulation.
\newblock In \emph{Proceedings of the 41st International Conference on Machine Learning (ICML)}, 2024{\natexlab{c}}.
\newblock URL \url{https://arxiv.org/abs/2311.01455}.

\bibitem[Zheng et~al.(2024)Zheng, Yan, Liu, Feng, Kang, and Ma]{robocas}
Zheng, L., Yan, F., Liu, F., Feng, C., Kang, Z., and Ma, L.
\newblock {RoboCAS}: A benchmark for robotic manipulation in complex object arrangement scenarios.
\newblock In \emph{NeurIPS Datasets and Benchmarks Track}, 2024.

\end{thebibliography}
\bibliographystyle{icml2025}

\newpage
\appendix
\onecolumn
\section{Appendix A: Pangu-7B Supervised Fine-Tuning for Motion Planning}
\label{appendix:pangu_sft}

In this section, we detail the Supervised Fine-Tuning (SFT) process of the Pangu-7B model, which is designed to synthesize motion planning code for sub-tasks within the V-CAGE task chain. Our goal is to enable the model to generate correct and robust low-level control policies based on high-level instructions.

\subsection{Data Collection and Augmentation}
To construct a high-quality SFT dataset, we leveraged expert data from the RoboTwin benchmark. Due to the limited number of original expert demonstrations, we employed a data augmentation strategy using the DeepSeek model.
Specifically, we started with 50 unique expert motion planning scripts from RoboTwin. We prompted DeepSeek to rewrite these scripts, generating functionally equivalent code with diverse logic structures and variable naming conventions while ensuring correctness. This data augmentation process resulted in a curated dataset of approximately 900 instruction-code pairs, which served as the foundation for our SFT stage.

\subsection{Fine-Tuning and Evaluation}
We fine-tuned the Pangu-7B base model using the generated dataset to adapt it to the specific API calls and physical constraints of the simulation environment.
To verify the effectiveness of our approach, we evaluated the fine-tuned model (Ours-Pangu-SFT) against the original pre-trained model (Pangu-Base) on a suite of RoboTwin validation tasks. For each task, we conducted 10 independent trials. We report two metrics: the Average Success Rate (Avg SR) across all trials, and the Top-3 Success Rate (Top3 SR), which reflects the model's peak performance capability.

\subsection{Results}
The quantitative results are presented in Table \ref{tab:pangu_sft_results}.
The Pangu-Base model failed to complete any of the tasks (0.0\% success rate), highlighting the difficulty of the motion planning format without specific adaptation. In contrast, the Pangu-SFT model demonstrated significant improvements, achieving high success rates on various complex manipulation tasks such as \texttt{place\_mouse\_pad} (100\% Avg SR) and \texttt{grab\_roller} (90\% Avg SR). The results confirm that fine-tuning on the DeepSeek-augmented expert data successfully enables Pangu-7B to master the intricate motion planning logic required for the V-CAGE tasks.

\begin{table}[h]
\centering
\caption{Comparison of Success Rates between Pangu-SFT and Pangu-Base on RoboTwin Tasks. 'Avg SR' denotes the average success rate over 10 runs, and 'Top3 SR' denotes the average of the top 3 runs.}
\label{tab:pangu_sft_results}
\resizebox{\columnwidth}{!}{%
\begin{tabular}{l|cc|cc}
\hline
\multirow{2}{*}{\textbf{Task Name}} & \multicolumn{2}{c|}{\textbf{Pangu-SFT}} & \multicolumn{2}{c}{\textbf{Pangu-Base}} \\
 & \textbf{Avg SR} & \textbf{Top3 SR} & \textbf{Avg SR} & \textbf{Top3 SR} \\
\hline
adjust\_bottle & 74.0\% & 96.7\% & 0.0\% & 0.0\% \\
beat\_block\_hammer & 59.0\% & 100.0\% & 0.0\% & 0.0\% \\
blocks\_ranking\_rgb & 60.0\% & 100.0\% & 0.0\% & 0.0\% \\
blocks\_ranking\_size & 12.0\% & 40.0\% & 0.0\% & 0.0\% \\
click\_alarmclock & 56.0\% & 90.0\% & 0.0\% & 0.0\% \\
click\_bell & 10.0\% & 33.3\% & 0.0\% & 0.0\% \\
grab\_roller & 90.0\% & 100.0\% & 0.0\% & 0.0\% \\
hanging\_mug & 6.0\% & 20.0\% & 0.0\% & 0.0\% \\
lift\_pot & 10.0\% & 33.3\% & 0.0\% & 0.0\% \\
move\_and\_press\_stapler & 9.0\% & 30.0\% & 0.0\% & 0.0\% \\
move\_can\_pot & 19.0\% & 60.0\% & 0.0\% & 0.0\% \\
move\_pillbottle\_pad & 7.0\% & 23.3\% & 0.0\% & 0.0\% \\
move\_playingcard\_away & 59.0\% & 100.0\% & 0.0\% & 0.0\% \\
move\_stapler\_pad & 65.0\% & 83.3\% & 0.0\% & 0.0\% \\
place\_a2b\_right & 45.0\% & 80.0\% & 0.0\% & 0.0\% \\
place\_bread\_basket & 3.0\% & 10.0\% & 0.0\% & 0.0\% \\
place\_dual\_shoes & 27.0\% & 73.3\% & 0.0\% & 0.0\% \\
place\_empty\_cup & 10.0\% & 33.3\% & 0.0\% & 0.0\% \\
place\_fan & 78.0\% & 90.0\% & 0.0\% & 0.0\% \\
place\_mouse\_pad & 100.0\% & 100.0\% & 0.0\% & 0.0\% \\
place\_object\_stand & 27.0\% & 90.0\% & 0.0\% & 0.0\% \\
place\_phone\_stand & 67.0\% & 96.7\% & 0.0\% & 0.0\% \\
press\_stapler & 63.0\% & 90.0\% & 0.0\% & 0.0\% \\
rotate\_qrcode & 85.0\% & 90.0\% & 0.0\% & 0.0\% \\
stack\_blocks\_two & 30.0\% & 100.0\% & 0.0\% & 0.0\% \\
stack\_bowls\_three & 18.0\% & 46.7\% & 0.0\% & 0.0\% \\
stamp\_seal & 27.0\% & 60.0\% & 0.0\% & 0.0\% \\
\hline
\end{tabular}%
}
\end{table}

\end{document}